\UseRawInputEncoding
\documentclass[letterpaper, 10 pt, conference]{ieeeconf}  % Comment this line out if you need a4paper

\IEEEoverridecommandlockouts                              % This command is only needed if 
\title{\LARGE \bf
Utilizing Navigation Paths to Generate Target Points for Enhanced End-to-End Autonomous Driving Planning
}

\author{Yuanhua Shen$^{1}$ Jun Li$^{2,*}$ % <-this % stops a space
\thanks{$^{1}$Yuanhua Shen is with University of Science and Technology of China, Hefei, China
        {\tt\small yuanhuashen@mail.ustc.edu.cn}}%
\thanks{$^{2}$Jun Li is with University of Science and Technology of China, Hefei, China
        {\tt\small Ljun@ustc.edu.cn}}%
\thanks{*Corresponding author: Jun Li}%
}
\usepackage{caption}
\usepackage{hyperref}
\usepackage{amssymb}
\usepackage{array}
\usepackage{booktabs}
\usepackage{multirow}
\usepackage{colortbl}
\usepackage{color,xcolor}
\usepackage{amsmath}
\usepackage{graphicx}
\begin{document}
\newsavebox\CBox
\def\textBF#1{\sbox\CBox{#1}\resizebox{\wd\CBox}{\ht\CBox}{\textbf{#1}}}

\maketitle
\thispagestyle{empty}
\pagestyle{empty}

%%%%%%%%%%%%%%%%%%%%%%%%%%%%%%%%%%%%%%%%%%%%%%%%%%%%%%%%%%%%%%%%%%%%%%%%%%%%%%%%
\begin{abstract}

In recent years, end-to-end autonomous driving frameworks have been shown to not only enhance perception performance but also improve planning capabilities. However, most previous end-to-end autonomous driving frameworks have focused primarily on enhancing environmental perception while neglecting the learning of autonomous vehicle driving intent, which refers to the vehicle's intended direction of travel. In planning, the autonomous vehicle's direction is clear and well-defined, yet this crucial aspect has often been overlooked. This paper introduces NTT (Navigation to Target for Trajectory planning), a method within an end-to-end framework for autonomous driving. NTT generates the planned trajectory in two steps. First, it generates the future target point for the autonomous vehicle on the basis of the navigation path. Then, it produces the complete planned trajectory on the basis of this target point. On the one hand, generating the target point for the autonomous vehicle from the navigation path enables the vehicle to learn a clear driving intent. On the other hand, generating the trajectory on the basis of the target point allows for a flexible planned trajectory that can adapt to complex environmental changes, thereby enhancing the safety of the planning process. Our method achieved excellent planning performance on the widely used nuScenes dataset and its effectiveness was validated through ablation experiments.

\end{abstract}

%%%%%%%%%%%%%%%%%%%%%%%%%%%%%%%%%%%%%%%%%%%%%%%%%%%%%%%%%%%%%%%%%%%%%%%%%%%%%%%%
\section{INTRODUCTION}

A robust autonomous driving system requires not only effective perception of the environment but also the ability to undertake rational and safe planning on the basis of environmental and navigation information.

Autonomous driving algorithms typically consist of several subtasks, including 3D object detection \cite{huang_huang_zhu_du,li_ge_yu_yang_wang_shi_sun_li_face++}, map segmentation \cite{Li_Wang_Wang_Zhao_2022,Liao_Chen_Wang_Cheng_Zhang_Liu_Huang_2022}, motion prediction \cite{Chai_Sapp_Bansal_Anguelov_2019,zhao2021tnt}, 3D occupancy prediction \cite{Tian_Jiang_Yun_Wang_Wang_Zhao_2023,tong2023scene} and planning \cite{Renz_Chitta_Mercea_Koepke_Akata_Geiger_2022,Cheng_Chen_Mei_Yang_Li_Liu_2023}. In recent years, an end-to-end approach \cite{Hu_Li_Wu_Li_Yan_Tao,hu2023planning} has integrated multiple independent tasks into multi-task learning, optimizing the entire system, including intermediate representations, toward the final planning task. UniAD \cite{hu2023planning} integrates six subtasks including object detection, object tracking, map segmentation, trajectory prediction, occupancy prediction, and planning into a unified end-to-end network framework for the first time. Compared with all previous methods, this approach achieves a comprehensive full-stack driving general model and improves the performance of all tasks.

\begin{figure}
\centering
\vspace{0.5cm}
\setlength{\abovecaptionskip}{0cm}
\hspace{-0.45cm}
\includegraphics[width=0.5\textwidth,height=0.25\textwidth]{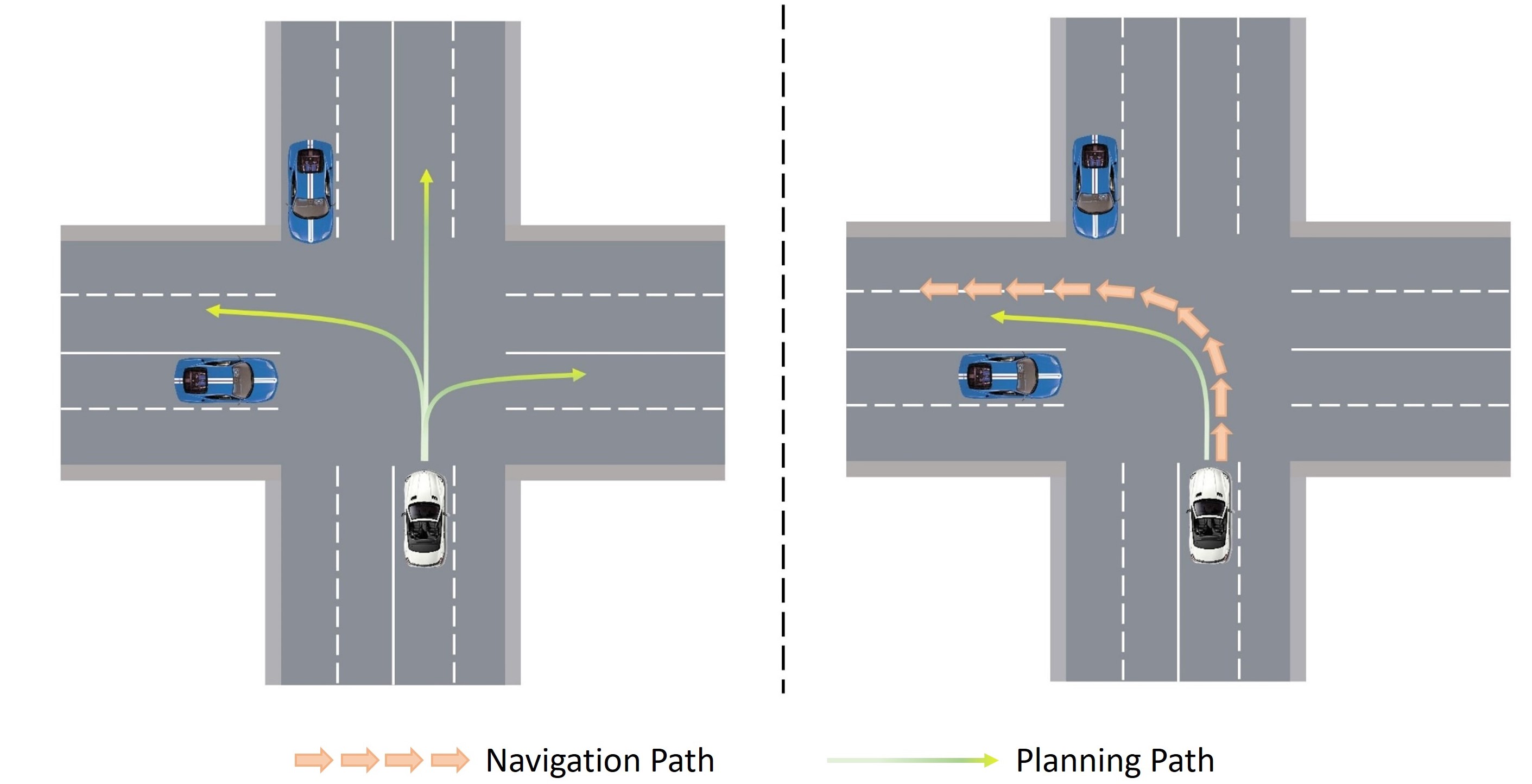}
\vspace{0cm}
\caption{Previous end-to-end planning approaches often relied on low-quality navigation information or treated it merely as a prediction task (left), leading to high uncertainty in the driving intent of planned trajectories. We advocate leveraging the navigation path to constrain planning (right), which results in more accurate driving intent for the ego vehicle.} \label{fig1}
\vspace{-0.5cm}
\end{figure}

However, most existing end-to-end methods underestimate the importance of navigation information in planning, often conflating planning with prediction into the same task. Learning-based planning and prediction algorithms are highly similar in their forms of representation. But they differ crucially in whether the driving intent, referring to the vehicle's intended direction of travel, is known. Prediction involves forecasting the future trajectories of agents on the basis of their current and past states, without knowledge of their intended direction of travel. The challenge in prediction lies in obtaining a highly uncertain multimodal distribution of future outcomes \cite{zhao2021tnt}. Planning requires generating a safe and comfortable trajectory for the ego vehicle, with its driving intent being explicit \cite{Hallgarten_Stoll_Zell_2023}. Some methods \cite{Hu_Li_Wu_Li_Yan_Tao,hu2023planning} attempt to constrain autonomous vehicle planning via discrete navigation commands (e.g., go straight, turn left, turn right). However, our experiments suggest that relying solely on simple navigation commands is insufficient for learning explicit driving intent. 

In this paper, we propose a method termed NTT (Navigation to Target for Trajectory planning), which utilizes the navigation path to constrain planning (as shown in Fig. \ref{fig1}), thereby clarifying the driving intent and enhancing planning performance. Specifically, NTT generates the planned trajectory in two steps. First, it uses the navigation path to generate a potential target point for the autonomous vehicle. This target point is then integrated with environmental information to derive the complete planning trajectory. This approach ensures a clear driving direction and allows for flexible planning that adapts to environmental changes, improving both correctness and safety. Our navigation paths are sourced from commercial software, which is readily available and practical. Due to the granularity limitations of such paths, there may be deviations between the path's starting point and the ego vehicle’s position. Consequently, we focus on the directional information of these paths. Through experiments, we have demonstrated the effectiveness of our method.

Our contributions can be summarized as follows.

\begin{itemize}
\item[$\bullet$]We introduce a planning methodology termed NTT that integrates navigation path data into an end-to-end framework. NTT employs a two-stage trajectory generation approach: it first generates a target point for the planned trajectory via the navigation path, and then generates the complete planned trajectory, ensuring that the planned trajectory is both safe and aligned with the navigation direction. Our method has been validated on the nuScenes \cite{Caesar_Bankiti_Lang_Vora_Liong_Xu_Krishnan_Pan_Baldan_Beijbom_2020} dataset.
\end{itemize}

\begin{itemize}
\item[$\bullet$]We enriched the nuScenes \cite{Caesar_Bankiti_Lang_Vora_Liong_Xu_Krishnan_Pan_Baldan_Beijbom_2020} dataset by incorporating navigation path information, offering valuable reference data for future research endeavors.
\end{itemize}

\section{RELATED WORK}
\subsection{Perception}
Perception forms the foundation of autonomous driving. In recent years, bird's-eye view (BEV) representation \cite{li_wang_li_xie_sima_lu_qiao_dai_2022} has emerged as a common strategy, enabling effective fusion of multimodal data and demonstrating significant potential across perception tasks, including 3D object detection \cite{huang_huang_zhu_du,li_ge_yu_yang_wang_shi_sun_li_face++,zhang_zhu_zheng_huang_huang_zhou_lu,Liu_Wang_Zhang_Sun}, map segmentation \cite{Li_Wang_Wang_Zhao_2022,Liao_Chen_Wang_Cheng_Zhang_Liu_Huang_2022,Liao_Chen_Zhang_Jiang_Zhang_Liu_Huang,yuan2024streammapnet}, and 3D occupancy prediction \cite{Tian_Jiang_Yun_Wang_Wang_Zhao_2023,tong2023scene,Wei_Zhao_Zheng_Zhu_Zhou_Lu}. In the realm of 3D object detection, DETR3D \cite{Wang_Guizilini_Zhang_Wang_Zhao_Solomon} utilizes 3D queries to index corresponding image features. In map segmentation, HDMapNet \cite{Li_Wang_Wang_Zhao_2022} integrates data from cameras and LiDAR sensors to predict vectorized map elements. MapTR \cite{Liao_Chen_Wang_Cheng_Zhang_Liu_Huang_2022} and MapTRv2 \cite{Liao_Chen_Zhang_Jiang_Zhang_Liu_Huang} model map elements as sets of points with a set of equivariant transformations, accurately describing the shape of map elements and stabilizing the learning process. StreamMapNet \cite{yuan2024streammapnet} employs multi-point attention and temporal information to increase the stability of large-scale, high-precision map reconstruction.

\subsection{Prediction}
Accurate prediction of the movements of traffic participants is crucial for ensuring the safety of planning. Some prediction methods utilize historical trajectories and HD maps as inputs \cite{zhao2021tnt,Liang_Yang_Hu_Chen_Liao_Feng_Urtasun_2020,Gao_Sun_Zhao_Shen_Anguelov_Li_Schmid_2020,Gu_Sun_Zhao_2021,Deo_Wolff_Beijbom_2021}. TNT \cite{zhao2021tnt} samples anchor points from the roadmap and generates trajectories based on these points. The trajectories are then scored, and non-maximum suppression (NMS) is employed to select the final trajectory set. DenseTNT \cite{Gu_Sun_Zhao_2021} improves upon TNT \cite{zhao2021tnt} by densely sampling points on the map and using a goal set predictor module to output multimodal prediction trajectories. Additionally, some prediction methods \cite{zhang_zhu_zheng_huang_huang_zhou_lu,Gu_Hu_Zhang_Chen_Wang_Wang_Zhao_2022} use agent and map features extracted from BEV features as inputs and employ attention networks \cite{Zhu_Su_Lu_Li_Wang_Dai_2020} to output predicted trajectories.

\subsection{Planning}
Traditional rule-based planners \cite{Bouchard_Sedwards_Czarnecki_2022,Dauner_Hallgarten_Geiger_Chitta_2023} have achieved significant progress, but their lack of generalizability remains a challenge. Learning-based planners \cite{Pini_Perone_Ahuja_Ferreira_Niendorf_Zagoruyko_2022,hu2023imitation} show immense potential due to their compatibility with end-to-end autonomous driving frameworks and their ability to improve performance through large-scale data. ST-P3 \cite{Hu_Li_Wu_Li_Yan_Tao}, as the first to propose an end-to-end autonomous driving framework based on surround-view cameras, takes multiple snapshots from surround-view camera images as inputs and sequentially processes them through perception, prediction, and planning modules to output the final planning path. UniAD \cite{hu2023planning} cleverly integrates multiple perception and prediction tasks, improving the performance of all tasks. VAD \cite{Jiang_Chen_Xu_Liao_Chen_Zhou_Zhang_Liu_Huang_Wang_2023} converts rasterized map representations into vectorized representations and further improves planning performance through three instance-level constraints. GenAD \cite{Zheng_Song_Guo_Chen_2024} models autonomous driving within a generative framework, simultaneously outputting prediction and planning results. However, these end-to-end planning methods do not effectively utilize navigation information. In this paper, we explore how to leverage navigation paths within an end-to-end framework to obtain planning trajectories with clear driving intentions.

\begin{figure*}
\centering
\vspace{0.2cm}
\setlength{\abovecaptionskip}{-0cm}
\includegraphics[width=1.0\textwidth,height=0.41\textwidth]{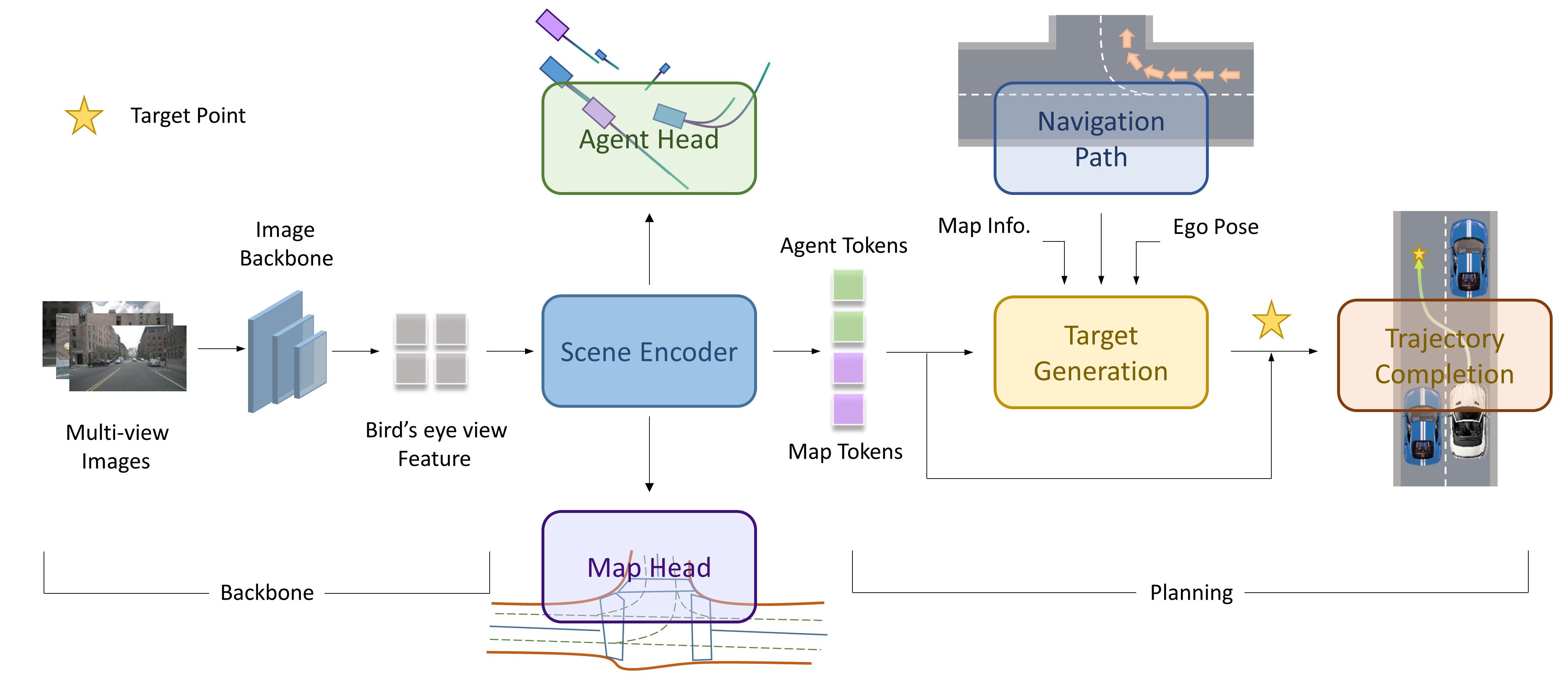}
\vspace{-0.2cm}
\caption{The NTT framework. Given surround-view images as inputs, an image feature network and a BEV encoder project the image features into the BEV feature space. The scene encoder learns scene tokens (including both agent and map tokens) from this space, which can be decoded into agent and map representations using respective heads. In the planning module, the planned trajectory is generated in two steps. First, the target generation module takes the navigation path, map information (decoded via the map head), ego pose, and scene tokens as inputs to produce the target point. The target point is then integrated with the scene tokens to generate the final planned trajectory.} \label{fig2}
\vspace{-0.3cm}
\end{figure*}

\section{METHODOLOGY}
This section details our NTT framework, illustrated in Fig. \ref{fig2}. NTT begins by using multiframe, multiview images as inputs and employs a convolutional network \cite{He_Zhang_Ren_Sun_2016} and a feature pyramid network \cite{Lin_Dollar_Girshick_He_Hariharan_Belongie_2017} to extract multi-scale image features. These features are then transformed into BEV representations via deformable cross-attention with BEV queries \cite{li_wang_li_xie_sima_lu_qiao_dai_2022}. Next, NTT utilizes map tokens and agent tokens to aggregate map and agent features from the BEV features, which can be decoded into specific vector representations (Sec. \ref{subsec:scene-encoder}). We then explain how navigation path data are integrated into the nuScenes \cite{Caesar_Bankiti_Lang_Vora_Liong_Xu_Krishnan_Pan_Baldan_Beijbom_2020} dataset and modeled (Sec. \ref{subsec:navigation-path}).Furthermore, we detail how these navigation data are used to generate the future target point for the ego vehicle and to derive the complete planning trajectory from this target point.(Sec. \ref{subsec:navigation-to-planning}). Finally, we discuss the training methodology for our end-to-end framework (Sec. \ref{subsec:training}).

\subsection{Scene Encoder}
\label{subsec:scene-encoder}
In autonomous driving scenarios, perceiving map elements and traffic participants is essential. Therefore, we use a set of map tokens $E_{M}$ and agent tokens $E_{A}$ to aggregate the map and agent features from the BEV features. Here, tokens refer to updatable embeddings. $E_{M}$ and $E_{A}$ are concatenated to form scene tokens $E_{s}$ represented as $E_{s} = \lbrack E_{M},E_{A} \rbrack$, which collectively describe the entire driving scenario.

{\bfseries BEV to map. } When each lane element is considered an instance, modeling the relationships between map elements and agents becomes straightforward. Thus, we employ a set of instance-level map tokens \cite{Liao_Chen_Wang_Cheng_Zhang_Liu_Huang_2022} $E_M$ to represent map features, where each map token can be decoded into a set of points in the BEV space along with corresponding class scores. In this paper, we consider four types of map elements (i.e., lane divides, road boundaries, pedestrian crossings and lane centerlines). Particularly for the lane centerline, according to LaneGap \cite{Liao_Chen_Jiang_Cheng_Zhang_Liu_Huang_Wang}, we use a complete and continuous path as the basic prediction unit. We use the decoded lane centerlines as one of the prior information sources for generating the target point. (Sec. \ref{subsec:navigation-to-planning} ).

{\bfseries BEV to agent. } Similarly, we use a set of instance-level agent tokens \cite{li_wang_li_xie_sima_lu_qiao_dai_2022} $E_A$ to represent agents. Through a 3D object detection head, each agent token can be decoded into the position, category scores, and heading angle. To further enrich the motion information of agent tokens, we employ attention mechanisms \cite{Vaswani_Shazeer_Parmar_Uszkoreit_Jones_Gomez_Kaiser_Polosukhin_2017} to enable interactions both among agents and between agents and the map. This facilitates the prediction of future multimodal trajectories for each agent, along with the corresponding probability scores for each trajectory modality.

\subsection{Navigation Path}
\label{subsec:navigation-path}
{\bfseries Navigational Path Acquisition. } The nuScenes \cite{Caesar_Bankiti_Lang_Vora_Liong_Xu_Krishnan_Pan_Baldan_Beijbom_2020} dataset inherently lacks navigation data, leading some research practices to construct navigation commands (e.g., go straight, turn left, turn right) on the basis of the ground truth trajectory of the ego vehicle. However, simplistic commands cannot provide specific and clear driving instructions for the ego vehicle. Thus, a more suitable approach involves guiding planning through the navigation path.

To further investigate whether end-to-end autonomous driving systems can achieve correct planning through the navigation path, we utilize the Google Maps API to obtain the navigation path for each scene in the nuScenes \cite{Caesar_Bankiti_Lang_Vora_Liong_Xu_Krishnan_Pan_Baldan_Beijbom_2020} dataset. Specifically, we first acquire the start and end points for each scene, and convert them to latitude and longitude coordinates on the basis of their map locations. These points are then used as inputs to obtain a rough navigation path through the Google Maps API. Next, we interpolate the obtained navigation path to fix the distance between each pair of navigation path points, setting the distance to 5 m in the experiments. The interpolated path serves as the final navigation path.

{\bfseries Navigation Path Model. } Given that only the future navigation path can be obtained during the actual driving process, we designate the nearest navigation path point to the current ego vehicle position as the starting point of the navigation route. The starting point, along with the subsequent $m$ points, forms the 2D navigation path group $P_{navi} \in \mathbb{R}^{(m + 1) \times 2}$ for the current frame.

Considering the meter-level accuracy of the navigation path obtained from the Google Maps API, which may exhibit significant positional deviation from the ground truth trajectory of the ego vehicle, our modeling focuses on the directional information of the navigation path. Specifically, we first calculate the difference between each pair of consecutive points in the path set to obtain the position vector $d_{i}$, which represents the relative positional relationship between every two points in the path set.
\begin{align}d_{i} = p_{i + 1} - p_{i}\end{align}
Here, $p_{i}$  represents the $i$-th point in the path group $P_{navi}$. Furthermore, to enhance the representation of directional features in the navigation path, we introduce the heading angle of the navigation path as an additional set of parameters. The heading angle can be calculated as the arctangent of the navigation position vector $d_{i} = \left( x_{i},y_{i} \right)$. The formula for the heading angle $h_{i}$ is as follows:
\begin{align} h_{i} = atan2\left( y_{i},x_{i} \right)\end{align}
In summary, we represent the navigation path as a set of vectors, where each vector $v_{i}$ can be considered a node. The node features are as follows:
\begin{align}v_{i} = \left\lbrack d_{i},{\cos{\left( h_{i} \right),}}{\sin\left( h_{i} \right)} \right\rbrack \end{align}
In this representation, since we will further extract features of the navigation path via neural networks, we normalize the heading angles via sine and cosine functions to avoid numerical overflow during the computation process. A navigation path $F_{navi}$ consists of $m$ nodes, denoted as $\{v_{1},v_{2},...,v_{m}\}$.

\subsection{Navigation to Planning}
\label{subsec:navigation-to-planning}
In this subsection, we introduce our two-step trajectory generation planning module. We detail how the target generation module uses prior information, such as the navigation path and map information decoded through the map head, combined with scene tokens, to generate the target point. We then explain how the target point interacts with scene tokens to obtain the final planning trajectory.

\begin{figure}
\centering
\vspace{0cm}
\setlength{\abovecaptionskip}{0cm}
\hspace{-0.45cm}
\includegraphics[width=0.5\textwidth,height=0.305\textwidth]{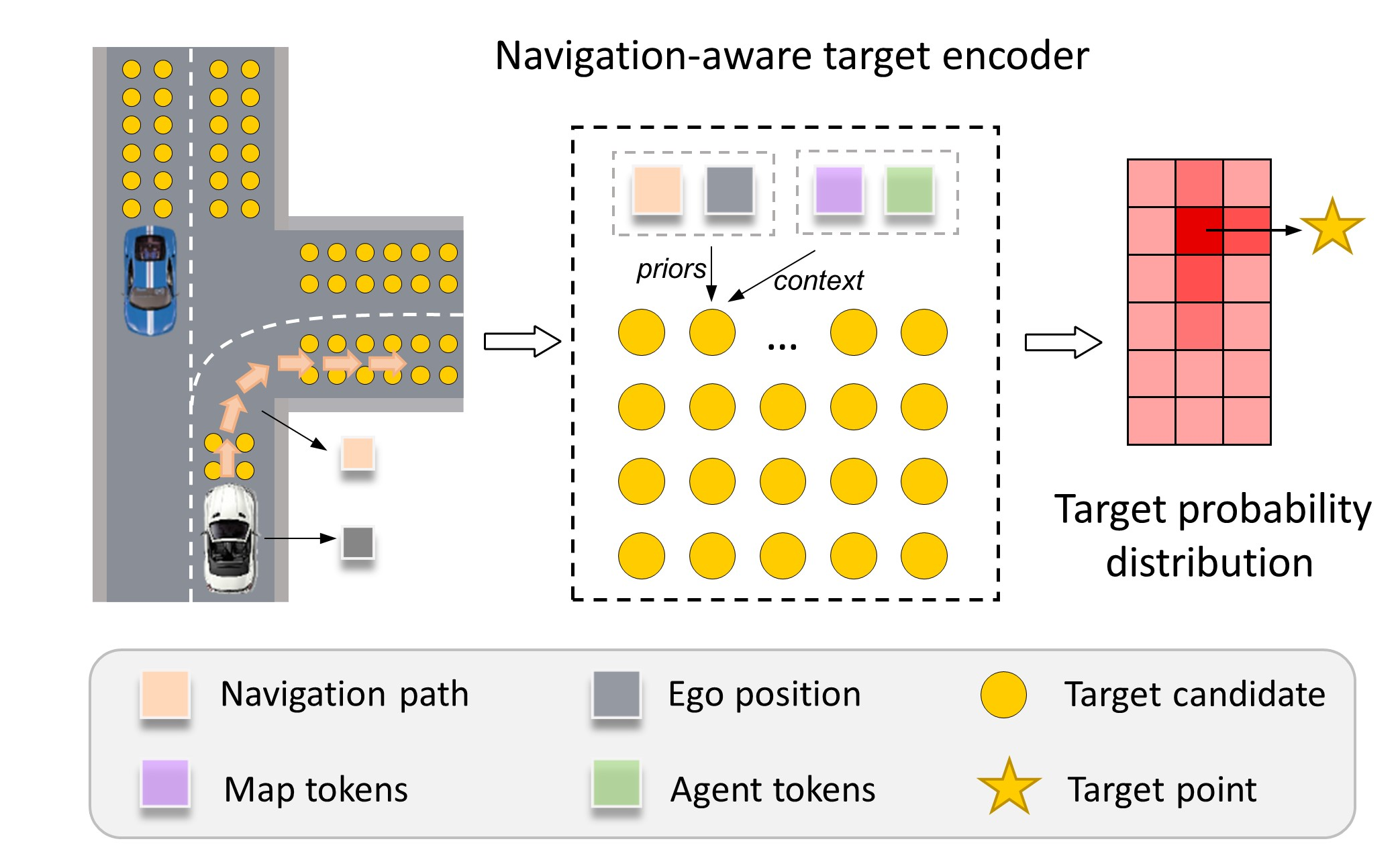}
\vspace{-0.2cm}
\caption{Description of the target generation module. In the navigation-aware target encoder, we use the navigation path as prior information and the scene information as context to learn the probability distribution of dense target candidates. The candidate point with the highest probability value is selected as the final target point.} \label{fig3}
\vspace{-0.3cm}
\end{figure}

{\bfseries Navigation to target. } 
Recent target-based methods \cite{zhao2021tnt,Zeng_Liang_Liao_Urtasun_2021,Gu_Sun_Zhao_2021} have shown excellent performance in prediction tasks. Methods such as TNT \cite{zhao2021tnt} and DenseTNT \cite{Gu_Sun_Zhao_2021} reduce the search space for target points and increase model robustness by presetting sampling points. We explore whether using the navigation path as prior information can further constrain the search space, enabling the model to learn the ego vehicle's explicit driving intent. This facilitates a natural transition from prediction to planning tasks. We illustrate the method for obtaining the target point based on sampling points in Fig. \ref{fig3}, which corresponds to the target generation module in Fig. \ref{fig2}.

Our method builds upon DenseTNT \cite{Gu_Sun_Zhao_2021}, with the key difference being the incorporation of navigation path information. Specifically, for the selection of sampling points, which serve as target candidates, we set dense points along the reconstructed lane centerlines and their surroundings. Since perceived lane centerlines may be inaccurate or unrecognized, we also set sampling points within a certain distance in front of the ego vehicle, ensuring forward movement even with unreliable perception results. We then use the navigation path as prior information to update the features of these target candidates, interact these features with the scene context to obtain new features, and stack the features across stages. Finally, we compute the probability distribution of the target candidates, with the highest-probability point being the final target point.

%%%%%%%%%%
%%%%%%%%%%%%%%%%%%%%

The initial representation of the navigation path features $F_{navi}$ is at the vector level, making fusing with the features of target candidates challenging. We use a two-layer VectorNet \cite{Liang_Yang_Hu_Chen_Liao_Feng_Urtasun_2020} encoding method to transform the navigation path features from the vector level to the instance level. A single-layer operation is illustrated as follows:
\begin{align} F_{navi}^{l+1} = Concat\left( g_{n} \left( F_{navi}^{l} \right),\varphi_{agg}\left( g_{n} \left( F_{navi}^{l} \right) \right) \right)\end{align}
where $F_{navi}^{l}$ represents the navigation path features at layer $l$, with $l=0$ as the initial navigation path representation $F_{navi}$. $g_{n}$ denotes a single-layer perceptron, and $\varphi_{agg}$ represents a max pooling operation. The features before and after max pooling are concatenated. The final instance-level navigation path features $P$ can be represented as:
\begin{align} P = \varphi_{agg}\left( F_{navi}^{2} \right) + g_{e}\left( p_{ego} \right) \end{align}
where $\varphi_{agg}$ denotes the max pooling operation applied to the navigation path features after two layers of VectorNet \cite{Liang_Yang_Hu_Chen_Liao_Feng_Urtasun_2020}  processing, $g_{e}$ is a single-layer perceptron, and $p_{ego} = \left( x_{ego},y_{ego} \right)$ represents the coordinates of the ego vehicle. The inclusion of the ego vehicle's positional information allows the model to incorporate spatial context directly, enhancing the relevance of the navigation path features to the vehicle's current location.

For the sampled target candidates $F_{t} \in \mathbb{R}^{N_{t} \times 2}$, which represent $N_{t}$ 2D coordinate points, we encode them via a single-layer perceptron $g_{1}$ to obtain the target candidate features $F$. Next, we concatenate $F$ with the navigation path features $P$, and then pass the concatenated features through another single-layer perceptron $g_{2}$ to obtain the integrated features $F_{1}$ that combine the navigation path and target candidate information.
\begin{align} 
F = g_{1}\left( F_{t} \right)
\end{align}
\begin{align} 
F_{1} = g_{2}\left( Concat \left( F, P \right) \right)
\end{align}

Subsequently, we further incorporate scene information $E_{s}=\lbrack E_{M},E_{A} \rbrack$, where $E_{M}$ represents the map tokens and $E_{A}$ represents the agent tokens. We use $F_{1}$ as the query and $E_{s}$ as the key and value in an attention mechanism \cite{Vaswani_Shazeer_Parmar_Uszkoreit_Jones_Gomez_Kaiser_Polosukhin_2017} to obtain the new integrated features $F_{2}$.

\begin{equation}
\begin{aligned} 
&F_{2} = CrossAttention\left( Q,K,V \right) \\
&Q = F_{1}, K = V = E_{s}
\end{aligned}
\end{equation}

Finally, we concatenate the navigation features $P$ with the features $F_{1}$ and $F_{2}$. The concatenated features are then passed through a single-layer perceptron $g_{3}$ for feature integration. A Softmax function is applied to compute the probability distribution over all target candidates.
\begin{align} 
Prob\left( F_{t} \right) = Softmax\left( g_{3} \left( Concat \left( P,F_{1},F_{2} \right) \right) \right)
\end{align}

We select the point with the highest probability as the target point $p_{t}$ for the autonomous vehicle. $p_{t}$ is a 2D coordinate represented as $\left( x_{t},y_{t} \right)$

{\bfseries Trajectory Completion. } The complete planned trajectory is generated based on the target point $p_{t}$ obtained from the target generation module. We encode the target point $p_{t}$ using a two-layer perceptron $g_{4}$ and then add it to the navigation path features $P$ to obtain the feature $Q_{ego}$, which represents the driving intent features of the ego vehicle.
\begin{align}
Q_{ego} = P + g_{4}\left( p_{t} \right)
\end{align}

Next, we use $Q_{ego}$ as the query and the scene tokens $E_{s}$ as the key and value in a cross-attention mechanism \cite{Vaswani_Shazeer_Parmar_Uszkoreit_Jones_Gomez_Kaiser_Polosukhin_2017} to obtain the updated feature $Q_{ego}^{'}$
\begin{equation}
\begin{aligned} 
&Q_{ego}^{'} = CrossAttention\left( Q,K,V \right) \\
&Q = Q_{ego}, K = V = E_{s}
\end{aligned}
\end{equation}

Finally, we concatenate the ego vehicle's driving intent feature $Q_{ego}$ with the scene-integrated feature $Q_{ego}^{'}$, and input them into a three-layer perceptron $g_{5}$ to decode the complete planned trajectory $\hat{T} \in \mathbb{R}^{k \times 2}$, where $k$ represents the future trajectory planned for k frames.
\begin{align}
\hat{T} = g_{5}\left(  Concat \left( Q_{ego},~Q_{ego}^{'} \right) \right)
\end{align}

\subsection{Training}
\label{subsec:training}
{\bfseries Scene Learning Loss. } We follow the design of VAD \cite{Jiang_Chen_Xu_Liao_Chen_Zhou_Zhang_Liu_Huang_Wang_2023}'s scene perception, dividing scene learning into two parts: learning vectorized maps and learning traffic participants' information. The map loss $\mathcal{L}_{map}$  consists of $l_{1}$ regression loss between the predicted map points and the ground truth map points, as well as focal loss \cite{Lin_Goyal_Girshick_He_Doll} for map classification. The loss for traffic participants $\mathcal{L}_{agent}$ includes 3D detection loss and motion prediction loss. The 3D detection loss consists of a classification loss based on focal loss \cite{Lin_Goyal_Girshick_He_Doll} and a bounding box regression loss based on $l_{1}$ loss. For motion prediction, we select the trajectory with the minimum final displacement error (minFDE) from the predicted $N_a$ trajectories to compute the $l_{1}$ loss with the ground truth trajectory as the regression loss, and use focal loss for the classification loss of the $N_{a}$ trajectories.

{\bfseries Target Probability Estimation Loss. } Following DenseTNT \cite{Gu_Sun_Zhao_2021}, we assign a score of 1 to the target candidate point closest to the endpoint of the ego vehicle's ground truth trajectory, and 0 to the rest. We compute the binary cross-entropy loss between the predicted target scores and the ground truth target scores as the target probability estimation loss $\mathcal{L}_{target}$.

{\bfseries Planning Loss. } Following the design of planning loss in VAD \cite{Jiang_Chen_Xu_Liao_Chen_Zhou_Zhang_Liu_Huang_Wang_2023}, we divide the planning loss $\mathcal{L}_{plan}$ into four parts. For the collision loss $\mathcal{L}_{col}$ between the ego vehicle and agents, a threshold $\alpha_{col}$ is set. We compute the distance $d_{a}^{i}$ between the ego vehicle and the nearest agent at each future timestamp. If $d_{a}^{i} < \alpha_{col}$, the collision loss is $\alpha_{col} - d_{a}^{i}$; otherwise, it is 0. $\mathcal{L}{col}$ is the sum of the collision losses over all future timestamps. For the collision loss $\mathcal{L}_{bd}$ between the ego vehicle and boundary, we compute the distance $d{b}^{i}$ between the ego vehicle and the nearest boundary line at each future timestamp. Similarly, if $d_{b}^{i} < \alpha_{bd}$, the collision loss is $\alpha_{bd} - d_{b}^{i}$; otherwise, it is 0, where $\alpha_{bd}$ is the set boundary threshold. The direction loss $\mathcal{L}_{dir}$ is the angular difference between the vector of the predicted trajectory of the ego vehicle and the vector of the nearest lane divider. The regression loss $\mathcal{L}_{reg}$ is the $l_{1}$ loss between the planned trajectory and the ground truth trajectory of the ego vehicle. The total planning loss can be formulated as the weighted average of the aforementioned losses:
\begin{align}\mathcal{L}_{plan} = \omega_{1}\mathcal{L}_{col} + 
\omega_{2}\mathcal{L}_{bd} +
~\omega_{3}\mathcal{L}_{dir} + ~\omega_{4}\mathcal{L}_{reg}\end{align}

where $\omega_{1}$, $\omega_{2}$, and $\omega_{3}$ are balancing parameters set to 1.0, 1.0, 0.5, and 1.0, respectively. The thresholds $\alpha_{col}$ and $\alpha_{bd}$ are set to 3 meters and 1 meter, respectively.

%%%%%%%%%%%%%%%%%%%%%%%%%%%%%%%
\begin{table*}[t]
% \large
% \normalsize
\small
\centering
\vspace{0.2cm}
\begin{tabular}{c | c| c c c c|c c c c }
\specialrule{0.5pt}{0pt}{0pt}
\toprule
\multirow{2}{*}{Method} & \multirow{2}{*}{Input} & \multicolumn{4}{c|}{L2(m)$\downarrow$}&\multicolumn{4}{c}{Collision Rate(\%)$\downarrow$}\\
% \cline{3-10}

 & & 1s & 2s & 3s & \cellcolor{gray!25}Avg. & 1s & 2s & 3s & \cellcolor{gray!25} Avg. \\
\midrule

FF \cite{Hu_Huang_Dolan_Held_Ramanan_2021}& Lidar &0.55	&1.20	&2.54	&\cellcolor{gray!25}1.43	&0.06	&0.17 &1.07 & \cellcolor{gray!25}0.43\\

EO \cite{Khurana_Hu_Dave_ZIglar_Held_Ramanan_2022}& Lidar &0.67	&1.36	&2.78	&\cellcolor{gray!25}1.60	&0.04	&0.09 &0.88 &\cellcolor{gray!25}0.33\\
\midrule

ST-P3 \cite{Hu_Li_Wu_Li_Yan_Tao}& camera &1.33	&2.11	&2.90	&\cellcolor{gray!25}2.11	&0.23	&0.62 &1.27 &\cellcolor{gray!25}0.71\\

UniAD \cite{hu2023planning}& camera &0.48	&0.96	&1.65	&\cellcolor{gray!25}1.03	&\textBF{0.05}	&\textBF{0.17} &0.71 &\cellcolor{gray!25}0.31\\

VAD-Tiny \cite{Jiang_Chen_Xu_Liao_Chen_Zhou_Zhang_Liu_Huang_Wang_2023}& camera &0.46	&0.76	&1.12	&\cellcolor{gray!25}0.78	&0.21	&0.35 &0.58 &\cellcolor{gray!25}0.38\\  

VAD-Base \cite{Jiang_Chen_Xu_Liao_Chen_Zhou_Zhang_Liu_Huang_Wang_2023}& camera &0.41	&0.70	&1.05 &\cellcolor{gray!25}0.72	&0.07	&\textBF{0.17} &0.41 &\cellcolor{gray!25}0.22\\

NTT& camera & \textBF{0.33}	&\textBF{0.58} &\textBF{0.93}	&\cellcolor{gray!25}\textBF{0.61}	&0.12	&0.18 &\textBF{0.29} &\cellcolor{gray!25}\textBF{0.20}\\ 

\midrule
\specialrule{0.5pt}{0pt}{1pt} 

\end{tabular}
\caption{Comparison with state-of-the-art methods on planning performance on the nuScenes \cite{Caesar_Bankiti_Lang_Vora_Liong_Xu_Krishnan_Pan_Baldan_Beijbom_2020} validation dataset. We assess autonomous driving performance via the average L2 error and collision rate within 1 s, 2 s, and 3 s. Lower values of the L2 error and collision rate indicate better performance. NTT demonstrates superior end-to-end planning performance.}
\label{table1}
\vspace{-0.3cm}
\end{table*}
%%%%%%%%%%%%%%%%%%%%%%%%%%%%%%%

{\bfseries Overall loss. } The total loss is the weighted sum of the losses from all the aforementioned modules, formulated as:
\begin{align}
\mathcal{L} = \omega_{m}\mathcal{L}_{map} + ~\omega_{a}\mathcal{L}_{agent} \notag \\ + ~~\omega_{t}\mathcal{L}_{target} + \omega_{p}\mathcal{L}_{plan}
\end{align}
where $\omega_{m}$, $\omega_{a}$, $\omega_{t}$, and $\omega_{p}$ are weighting coefficients. We use a two-stage training approach for NTT. First, we train the perception and prediction modules. Then, we add the planning module to achieve end-to-end optimization of all modules. To exclude a module from the parameter updates, we set the coefficient for the corresponding loss to 0. In the first stage, we set $\omega_{m}$ to 1.0, $\omega_{a}$ to 0.25, and both $\omega_{t}$ and $\omega_{p}$ to 0. In the second stage, we set $\omega_{m}$, $\omega_{a}$, $\omega_{t}$, and $\omega_{p}$ to 1.0, 0.25, 0.2, and 1.0, respectively.

%%%%%%%%%%%%%%%%%%%%%%%%%
\begin{table}[t]
% \tiny
% \small
% \normalsize
% \large
\centering
\begin{tabular}{c | c c c c|c c c c }
\specialrule{0.5pt}{0pt}{0pt}
\toprule
\multirow{2}{*}{Target} & \multicolumn{4}{c|}{L2(m)$\downarrow$}&\multicolumn{4}{c}{Collision Rate(\%)$\downarrow$}\\
% \cline{3-10}

  & 1s & 2s & 3s & \cellcolor{gray!25}Avg. & 1s & 2s & 3s & \cellcolor{gray!25} Avg. \\
\midrule

- & 0.35	&0.60	&0.93	&\cellcolor{gray!25}0.63	&0.56	&0.70 &0.88 & \cellcolor{gray!25}0.71\\

\checkmark & \textBF{0.33}	&\textBF{0.58} &\textBF{0.93}	&\cellcolor{gray!25}\textBF{0.61}	&\textBF{0.12}	&\textBF{0.18} &\textBF{0.29} &\cellcolor{gray!25}\textBF{0.20}\\ 

\midrule
\specialrule{0.5pt}{0pt}{1pt} 

\end{tabular}
\caption{An ablation study to validate the importance of the target generation module.}
\label{table2}
\vspace{-0.5cm}
\end{table}
%%%%%%%%%%%%%%%%%%%%

\section{Experiments}

\subsection{Dataset and Metric}
We evaluate our proposed method on the popular nuScenes \cite{Caesar_Bankiti_Lang_Vora_Liong_Xu_Krishnan_Pan_Baldan_Beijbom_2020} dataset, which contains 1000 driving scenes, each lasting roughly 20 seconds. The dataset includes 700 scenes for training, 150 for validation, and 150 for testing. Keyframe annotations are provided at 2 Hz. Each sample includes RGB images from 6 cameras, covering a $360^\circ$ horizontal field of view of the ego vehicle. Following existing end-to-end autonomous driving methods \cite{Hu_Li_Wu_Li_Yan_Tao,hu2023planning,Jiang_Chen_Xu_Liao_Chen_Zhou_Zhang_Liu_Huang_Wang_2023}, we use the L2 displacement error and collision rate to measure the quality of planning. The L2 displacement error measures the L2 distance between the planning trajectory and the ground truth trajectory. The collision rate measures the frequency of collisions with other traffic participants under the planning trajectory. We use the past 2 seconds of information as input and evaluate the planning performance for the next 3 seconds.

\subsection{Implementation Details}
We adopt ResNet50 \cite{He_Zhang_Ren_Sun_2016} as the backbone network to extract image features, with the input images resized to $640 \times 360$. We employ $200 \times 200$ BEV tokens to perceive the driving scene within a range of $60 m \times 30 m$. We set the number of map tokens $E_{M}$ to 100 and the number of agent tokens $E_{A}$ to 300. Each map token further comprises 20 point tokens to represent map points. We set the number of navigation path points $m$ to 10, meaning that the autonomous vehicle receives 50 meters of navigation path at a time. Given that the majority of the scenes in the nuScenes dataset are urban, this distance is sufficient. For each detected object, we set the number of predicted trajectories $N_{a}$ to 6. For the ego vehicle, we output only one planned trajectory for the next 3 seconds, with one point every 0.5 seconds, resulting in 6 future planning frames ($k=6$). For training, we employ the AdamW \cite{Loshchilov_Hutter_2017} optimizer with a cosine annealing \cite{Loshchilov_Hutter_2016} scheduler. The initial learning rate is set to $1 \times 10^{- 4}$ with a weight decay of 0.01. We trained for 60 epochs on 4 NVIDIA Quadro RTX 6000 GPUs, with a total batch size of 4.

\subsection{Main Results}
We compared NTT with state-of-the-art end-to-end driving methods in Tab. \ref{table1}. It is evident that NTT achieves state-of-the-art performance. Compared with VAD-Base \cite{Jiang_Chen_Xu_Liao_Chen_Zhou_Zhang_Liu_Huang_Wang_2023}, NTT reduces planning displacement errors within 1 s, 2 s, and 3 s, with an average decrease of 0.11 m across all three intervals, from 0.72 m to 0.61 m. Additionally, NTT has the lowest collision rate within 3 s compared with the other methods in the table. Specifically, we reduced the collision rate by 0.12\% compared with VAD-Base, from 0.41\% to 0.29\%. These results demonstrate the effectiveness of NTT in improving both safety and accuracy in end-to-end planning.

%%%%%%%%%%%%%%%%%%%%%%%%%
\begin{table}[t]
% \tiny
% \small
% \normalsize
% \large
\centering
\begin{tabular}{c | c c c c|c c c c }
\specialrule{0.5pt}{0pt}{0pt}
\toprule
\multirow{2}{*}{Setting} & \multicolumn{4}{c|}{L2(m)$\downarrow$}&\multicolumn{4}{c}{Collision Rate(\%)$\downarrow$}\\
% \cline{3-10}

  & 1s & 2s & 3s & \cellcolor{gray!25}Avg. & 1s & 2s & 3s & \cellcolor{gray!25} Avg. \\
\midrule

VAD  & 1.09	&1.84	&2.70	&\cellcolor{gray!25}1.88	&0.0	&0.84 &2.43 & \cellcolor{gray!25}1.09\\

tgt+emb & 0.77	&1.45 &2.34	&\cellcolor{gray!25}1.52	&0.0 &0.84&1.50&\cellcolor{gray!25}0.78\\ 

tgt+cmd & 1.00	&1.79 &2.79	&\cellcolor{gray!25}1.86	&0.0&0.84&2.99&\cellcolor{gray!25}1.28\\ 

tgt+path & \textBF{0.76}	&\textBF{1.36} &\textBF{2.09}	&\cellcolor{gray!25}\textBF{1.40}	&\textBF{0.0}	&\textBF{0.28} &\textBF{1.12} &\cellcolor{gray!25}\textBF{0.47}\\ 

\midrule
\specialrule{0.5pt}{0pt}{1pt} 

\end{tabular}
\caption{We selected several turning scenarios from the nuScenes \cite{Caesar_Bankiti_Lang_Vora_Liong_Xu_Krishnan_Pan_Baldan_Beijbom_2020} validation set and validated the effectiveness of NTT in utilizing the navigation path in these turning scenarios. "tgt+path" represents NTT, which utilizes the navigation path as prior knowledge for the target point generation.}
\label{table3}
\vspace{-0.5cm}
\end{table}
%%%%%%%%%%%%%%%%%%%%

\begin{figure*}
\centering
\vspace{0.2cm}
\setlength{\abovecaptionskip}{-0cm}
\includegraphics[width=1.0\textwidth,height=0.25\textwidth]{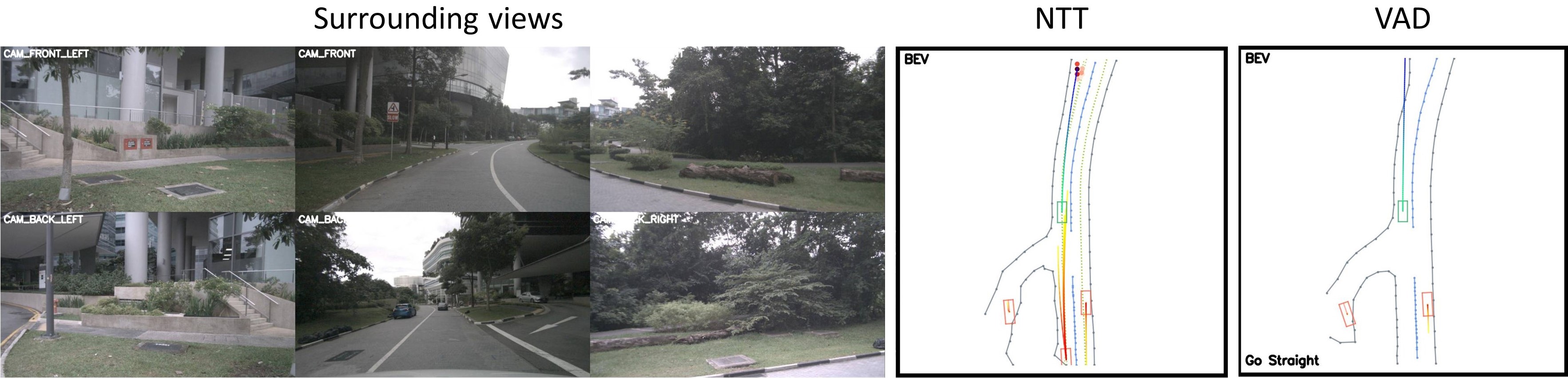}
\includegraphics[width=1.0\textwidth,height=0.23\textwidth]{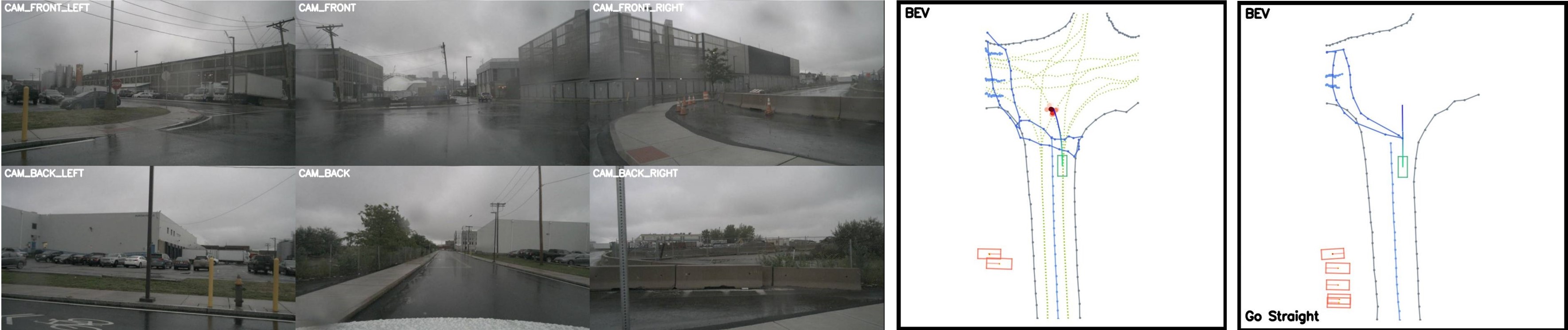}
\includegraphics[width=1.0\textwidth,height=0.23\textwidth]{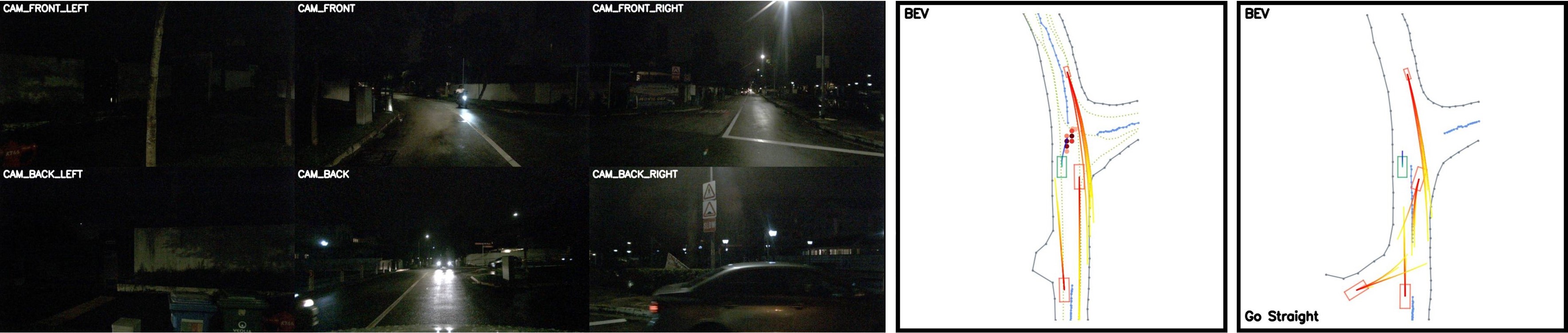}
\vspace{-0.2cm}
\caption{Visual comparison of NTT and VAD \cite{Jiang_Chen_Xu_Liao_Chen_Zhou_Zhang_Liu_Huang_Wang_2023}. We provide comparative visualizations for daytime, cloudy, and nighttime conditions, along with visual results for perception, prediction, and planning.} \label{fig4}
\end{figure*}

\subsection{Ablation Study}
\label{subsec:ablation}
{\bfseries Significance of the target generation module. } We conducted an ablation study to verify the effectiveness of the target generation module in Tab. \ref{table2}. We attempt to remove the module and directly use the encoded features of the navigation path as the query. The query interacts with the scene tokens $E_{s}$ through an attention mechanism, and the updated features are then fed into a multi-layer perceptron to directly generate the final planned trajectory. As shown in Tab. \ref{table2}, when only the navigation path was used without the target generation module, there was a significant increase in the collision rates, and a certain degree of increase in the displacement error of the planning trajectory. Through the generated target point, a flexible planning trajectory can be obtained on the basis of the environment while keeping the target position unchanged. The experimental results demonstrate that this module effectively reduces the collision rate and improves accuracy to a certain degree.

{\bfseries Role of the navigation path. } We also analyzed the effectiveness of NTT in utilizing the navigation path for end-to-end planning. The quality of driving intent learning is reflected primarily in performance during turns, whereas planning during straight driving relies more on environmental perception for obstacle avoidance. Therefore, we conducted experiments in several turning scenarios. Specifically, we evaluated scenes from the nuScenes \cite{Caesar_Bankiti_Lang_Vora_Liong_Xu_Krishnan_Pan_Baldan_Beijbom_2020} validation set where the lateral movement of the ego vehicle's ground truth trajectory exceeded 2 meters within the next 3 seconds. In Tab. \ref{table3}, "tgt+emb" indicates the absence of using the navigation path as prior knowledge for generating the target point. Instead, the target point is directly generated from the driving environment through a learnable embedding. "tgt+cmd" represents the use of previous end-to-end methods (i.e., using navigation commands) to select the target point. "tgt+path" refers to NTT, which utilizes the navigation path as prior knowledge for target point generation. Compared with VAD \cite{Jiang_Chen_Xu_Liao_Chen_Zhou_Zhang_Liu_Huang_Wang_2023}, NTT significantly improves planning performance in turning scenarios, we observe a reduction of 0.48 m in the average planning displacement error and a 57\% decrease in collision rates. The comparison between the "tgt+path" and "tgt+cmd" results highlights the superiority of navigation paths over navigation commands, whereas the comparison between "tgt+cmd" and "tgt+emb." further emphasized the disadvantages of navigation commands, which may even have a negative effect on driving intent learning.

{\bfseries Visualizations. } We provide visual results of NTT and compare them with VAD \cite{Jiang_Chen_Xu_Liao_Chen_Zhou_Zhang_Liu_Huang_Wang_2023}. We adopt the visualization approach of VAD \cite{Jiang_Chen_Xu_Liao_Chen_Zhou_Zhang_Liu_Huang_Wang_2023}, showcasing mapping, detection, motion prediction, and planning results from a bird's-eye view perspective. Comparisons are made under different weather conditions: sunny, cloudy, and nighttime. In the top-down view shown in the right half of Fig. \ref{fig4}, the green rectangles represent the ego vehicle, and the red rectangles indicate detected agents. Each detected agent has multiple predicted trajectories associated with it. The blue-to-green gradient lines represent the planned trajectories. The gray polylines denote boundary lines, and blue polylines represent lane dividers. In the NTT visualization, compared with the VAD method, we have additionally included yellow dashed lines to denote the lane centerlines. The small red circles around the end points of the planned trajectories represent target candidates, with darker colors indicating higher probabilities of being the driving target. Here, we visualize only the top 10 points with the highest probability. The results of the planned trajectories clearly indicate that NTT performs better than VAD \cite{Jiang_Chen_Xu_Liao_Chen_Zhou_Zhang_Liu_Huang_Wang_2023} in capturing subtle turning intentions. Additionally, owing to the constraints of the target points, the planned trajectories better align with road structures, leading to more accurate and safer trajectory planning.

\section{CONCLUSIONS}

In this paper, we introduce NTT (Navigation to Target for Trajectory planning), a planning method that integrates navigation path information within an end-to-end framework. We explore a design that first utilizes the navigation path to constrain the generation of the target point, followed by generating the complete planning trajectory on the basis of the target point. Extensive experiments on the nuScenes \cite{Caesar_Bankiti_Lang_Vora_Liong_Xu_Krishnan_Pan_Baldan_Beijbom_2020} dataset demonstrate the superior planning performance of our proposed approach, NTT. In the future, further exploration of how to deploy end-to-end autonomous driving frameworks in real vehicles and achieve safe and efficient point-to-point planning is warranted..

% %%%%%%%%%%%%%%%%%%%%%%%%%

% \addtolength{\textheight}{-12cm}  
% This command serves to balance the column lengths
                                  % on the last page of the document manually. It shortens
                                  % the textheight of the last page by a suitable amount.
                                  % This command does not take effect until the next page
                                  % so it should come on the page before the last. Make
                                  % sure that you do not shorten the textheight too much.

% \end{thebibliography}

\bibliographystyle{IEEEtran}
\bibliography{IEEEabrv,mycite}
% \bibliography{mycite}

\end{document}